\documentclass[letterpaper, 10 pt, conference]{ieeeconf}  

\IEEEoverridecommandlockouts                              

\usepackage{mathptmx}       
\usepackage{helvet}         
\usepackage{courier}        
\usepackage{type1cm}        
\usepackage{makeidx}         
\usepackage{graphicx}        
\usepackage{multicol}        
\usepackage[bottom]{footmisc}
\usepackage{amsmath} 
\usepackage{amssymb}  
\usepackage{subfigure}
\usepackage{mdwlist}
\usepackage{changepage}
\usepackage{dblfloatfix}
\usepackage{algorithm}
\usepackage{algpseudocode}

\usepackage{epsfig} 

\title{\LARGE \bf Localizing Grasp Affordances in 3-D Points Clouds Using Taubin Quadric Fitting}

\author{Andreas ten Pas and Robert Platt
\thanks{The authors are with the College of Computer and Information Science at Northeastern University.
{\tt\small atp@ccs.neu.edu}, {\tt\small rplatt@ccs.neu.edu}}}

\begin{document}

\maketitle
\thispagestyle{empty}
\pagestyle{empty}

\begin{abstract}
Perception-for-grasping is a challenging problem in
robotics. Inexpensive range sensors such as the Microsoft Kinect
provide sensing capabilities that have given new life to the effort of
developing robust and accurate perception methods for robot
grasping. This paper proposes a new approach to localizing enveloping
grasp affordances in 3-D point clouds efficiently. The approach is
based on modeling enveloping grasp affordances as a cylindrical shells
that corresponds to the geometry of the robot hand. A fast and
accurate fitting method for quadratic surfaces is the core of our
approach. An evaluation on a set of cluttered environments shows high
precision and recall statistics. Our results also show that the
approach compares favorably with some alternatives, and that it is
efficient enough to be employed for robot grasping in real-time.

\end{abstract}

\section{Introduction}

Recently, the development of inexpensive range sensing technology such
as the Microsoft Kinect has given new life to the effort to develop
robust and accurate perceptual capabilities for robot
grasping. Perception-for-grasping is a challenging problem because
even small localization errors can cause the robot hand to miss the
target, resulting in complete grasp failure. One approach to the
problem is to attempt to localize all relevant objects in the
scene. This can be accomplished by creating a library of object
models~\cite{goldfeder_iros2009,brook_icra2011} that contains one
model for every object that might need to be grasped. The scene is
searched for objects from the library. When a match is found, a
manipulation planner decides how to pick up the target object. This
method is potentially very robust because it leverages prior
information about object geometry, but there are drawbacks. Building
and maintaining a suitable library is potentially very challenging and
performing the matching process can be computationally
expensive. Moreover, the method does not work at all for deformable
objects or in natural or unstructured environments where it is
impossible to predict object geometry in advance.

An alternative is to localize grasp geometries directly. For example,
rather than localizing a particular coffee mug found in a large
database and creating a plan to grasp it by the handle, the system
might localize the handle directly based on a prior knowledge of what
kinds of geometries the robot is capable of grasping. This corresponds
with the notion of a {\em grasp affordance}: a geometric
characteristic of an object that allows it to be grasped by a
particular robot hand or gripper~\footnote{The term ``affordance'' was
  originally used by Gibson~\cite{gibson_1977} to describe a
  characteristic of an object that enables a particular action to be
  performed with it.}. This approach has several potential
advantages. First, it is very flexible because there is no need to
create the object database and it can be applied to flexible or
unmodelled objects. Moreover, it has the potential to simplify
grasping because there is no need to do grasp planning. Each localized
grasp affordance corresponds directly to a set of hand poses to which
the robot can reach and achieve a grasp. In addition, it separates the
geometry of grasping from the semantic process of deciding what to do
with the object or how to grasp it (which affordance to use).

\begin{figure}
\begin{center}
    \subfigure[Enveloping grasp affordances circled in cyan]{\includegraphics[height=1.5in]{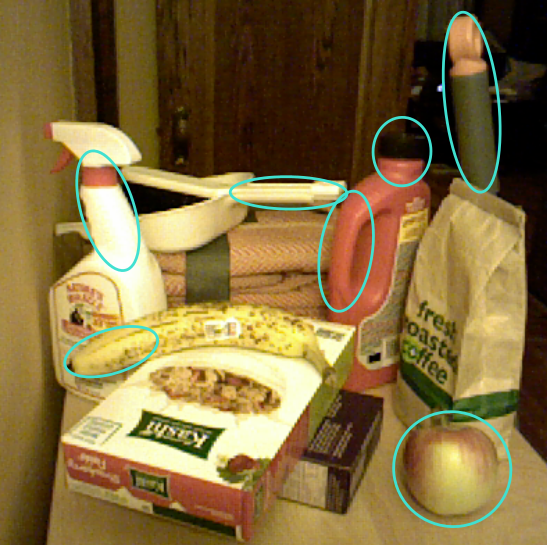}}
    \subfigure[Enveloping grasp affordances localized by our algorithm]{\includegraphics[height=1.5in]{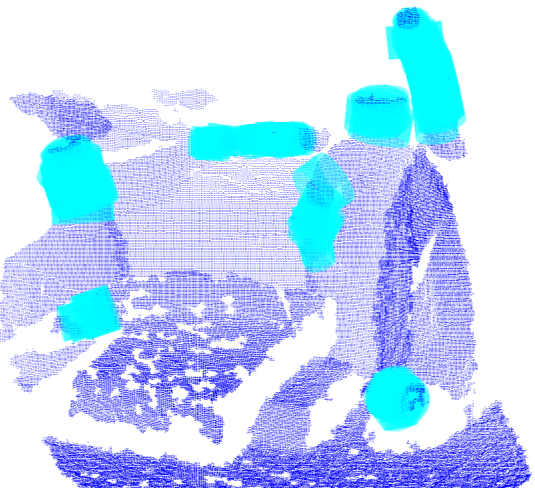}}
\end{center}
\caption{Illustration of enveloping grasp affordance localization. The
  objective is to precisely localize each location where an object can
  be securely grasped using an enveloping grasp. (b) shows grasp
  affordances identified by our algorithm. Notice that each of these
  affordances can be grasped by the PR2 robot hand illustrated in
  Figure~\ref{fig:affordance_illustration}.}
\label{fig:front_example}
\end{figure}

This paper proposes a new approach to localizing enveloping grasp
affordances in 3-D point clouds efficiently. The approach is based on
modeling an enveloping grasp affordance as a cylindrical shell that
corresponds to the geometry of the robot hand. The surface of the grasp
affordance must be contained inside the innermost radius of the shell
which must be no larger than the maximum hand aperture. The gap
between the inner and outer radii must be empty and sufficiently thick
to allow clearance for a robot hand to reach a grasping
configuration. We propose a perception pipeline that localizes these
cylindrical shells efficiently. The core of our approach is an
application of Taubin quadric fitting~\cite{taubin_pami1991} that
makes our algorithm faster and more accurate than alternative
methods. Our approach does not depend upon making any assumptions
about object separation or ground support planes. Our results indicate
that the approach is works well in cluttered environments such as that
illustrated in Figure~\ref{fig:front_example}. We show high precision
and recall statistics and show that the approach compares favorably
with some alternatives.

\subsection{Background}

Recently, there has been a strong interest in applying the bevy of new
range sensing technologies to the problem of
perception-for-grasping. One approach is to focus on localizing
modeled objects in the scene. After localizing an object with known
geometry, a grasp planner can be used to find a suitable grasp. Here,
it is increasingly common to use a feature-based approach to
localization. Two recent feature representations for use with 3-D
point clouds are Fast Point Feature Histograms
(FPFH)~\cite{rusu_icra2009} and the SHOT
feature~\cite{tombari_eccv2010}. Both of these encode the local feature
geometry in terms of a neighborhood of point locations and
normals. After choosing a feature representation, the next step is to
align features in the model with features found in the scene. One
well-known way to do this is to use iterative closest point
(ICP)~\cite{besl_pami92}. More recently, a stochastic generalization
of ICP was proposed~\cite{glover_iros2013}. Another popular approach
is Hough voting~\cite{tombari_psivt2010,sun_eccv2010}. These
strategies often require significant pre-processing of the point
cloud: voxelization, ground plane extraction, surface normals
estimation, {\em etc.} State of the art approaches can be expected to
achieve precision and recall results of between 70\% and 90\% for
cluttered and occluded scenes~\cite{glover_iros2013}.

From a practical perspective, it is often the case that not all
objects in a given scene can be identified with a modeled object from
a database. Recently, a growing body of work has focused on localizing
and modeling unknown objects. Some approaches work by representing
unknown objects using shape primitives. For example, Rusu {\em
  et. al.} represent kitchen environments using planes, boxes,
cylinders, {\em etc.}~\cite{rusu2008ras} and Biegelbauer and Vincze
describe complex shapes by fitting
superquadrics~\cite{biegelbauer_icra2007}. Other work includes
strategies for modeling and grasping unknown
objects~\cite{rusu_iros2009,brook_icra2011}. These strategies often
make strong assumptions about ground support planes and object
separation in order to make object segmentation easier.

Recent work has also focused on learning to detect graspable
geometries in a scene. For example, Jiang {\em et. al.} propose a
learning approach that predicts the ``graspability'' of parts of an
object~\cite{jiang_icra2011}. A potential grasp is represented as a
rectangle and a feature representation is proposed that enables a
classifier to achieve good prediction performance. Recently, related
work has achieved similar goals using an unsupervised
approach~\cite{lenz_rss2013}. While the objectives of the above work
are similar to our own, our current work is based on geometric
modeling rather than learning. In fact, it is notable and a bit
surprising that our work achieves such good performance (see
Section~\ref{sect:experiments}) without learning. Other recent work
closely related to our own has focused on localizing and learning to
localize grasp affordances in point
clouds~\cite{popovic_ras2010,barck_icar2009}.

\section{Grasp Affordance Geometry}
\label{sect:approach}

\begin{figure}
\begin{center}
    \subfigure[Hand Geometry]{\includegraphics[height=1.5in]{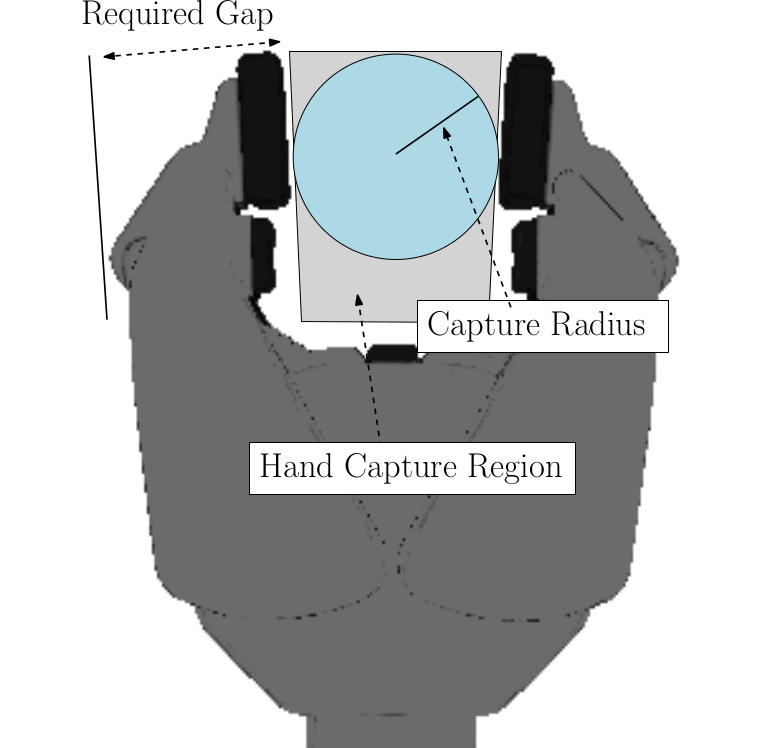}}
    \subfigure[EGA Geometry]{\includegraphics[height=1.7in]{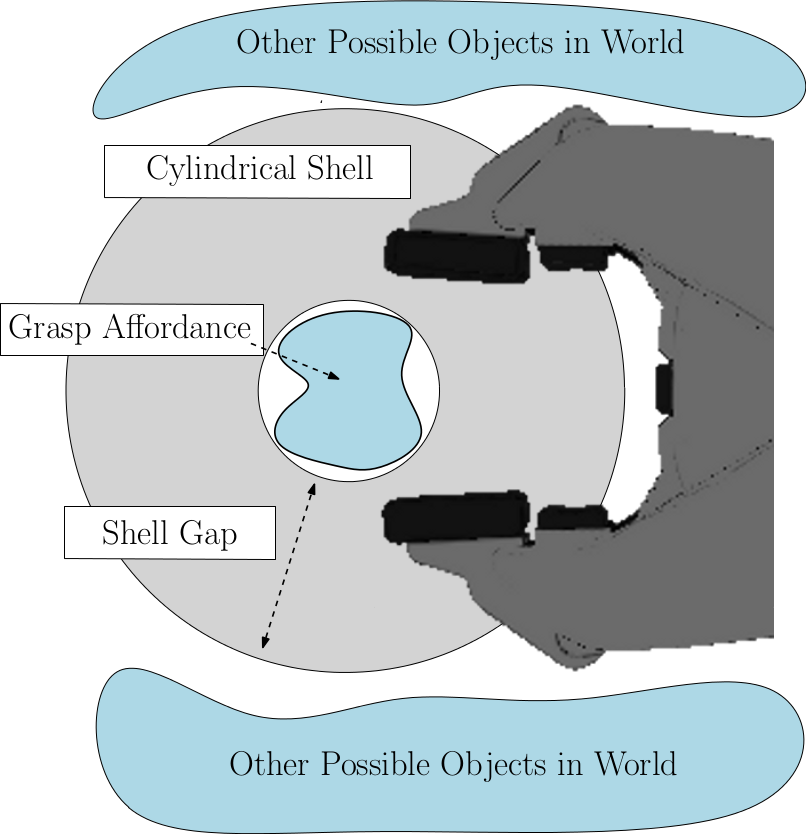}}
\end{center}
\caption{Illustration of (a) the hand geometry, and (b) the enveloping
  grasp affordance (EGA) geometry.}
\label{fig:affordance_illustration}
\end{figure}

We derive the geometry of an enveloping grasp affordance from the
robot hand geometry (see Figure~\ref{fig:affordance_illustration}). In
an enveloping grasp~\cite{cutkosky_wright_icra86}, the ``thumb''
opposes one or more ``fingers'' and the hand wraps most or all of the
way around the object. The plane in which this opposition occurs is
the {\em opposition plane}~\cite{iberall_book}. The axis perpendicular
to this plan is the {\em opposition axis}~\cite{iberall_book}. The
space in the opposition plane contained between the thumb and fingers
will be known as the {\em capture region}~\cite{dogar_rss11}. The
radius of the largest inscribed circle in the capture region will be
known as the {\em capture radius}. The maximum thickness of the thumb
or fingers in the opposition plane will be known as the {\em finger
  thickness}.

In order for an object in the environment to be grasped, two
conditions must be met: 1) a portion of the object surface must fit
within the capture region of the robot hand, and 2) the object must be
partially surrounded by a gap comprised of sufficient free space to
allow the gripper to pass. These two conditions can roughly be
translated into the following geometric charcaterization of a point
neighborhood in the 3-D cloud:
\begin{enumerate}
\item points that lie on the object surface must be contained within a
  cylinder with radius no larger than the capture radius;
\item this cylinder must be contained within a cylindrical shell that
  is clear of points and at least as thick as the finger thickness.
\end{enumerate}
These two conditions on a point neighborhood will be referred to as
the enveloping grasp affordance (EGA) conditions. Notice that the EGA
geometry is parameterized by the characteristics of the robot
hand. Any point cloud neighborhood satisfying the EGA conditions for
the given hand must be graspable in the sense that it is possible for
the robot hand to close around whatever object material is contained
within the shell.

\section{Localizing Candidate Grasp Affordances Using Taubin Fitting}
\label{sect:taubinfit}

Localizing environmental geometries that satisfy even the simple EGA
conditions can be challenging. A key idea in this paper is to sample a
large set of local neighborhoods from the point cloud and to use
Taubin quadric fitting to see if they contain any potential EGA
geometries. For each neighborhood, Taubin fitting is used to calculate
a smooth approximation of the local surface(s) efficiently. Then, we
evaluate whether the surface is likely to lie within the capture
radius of the hand by thresholding on median curvature and normal
covariance.

\subsection{Taubin Quadric Fitting}

Taubin fitting approximates the least-squares fit of a quadratic
surface in three variables to a set of points in Cartesian space. A
quadric can be described in implicit form by $f(\mathbf{c},\mathbf{x})
= 0$, where 
\begin{align} \nonumber
f(\mathbf{c},\mathbf{x}) = & c_1x_1^2 + c_2x_2^2 + c_3x_3^2 + c_4x_1x_2 + c_5x_2x_3 + \\
& c_6x_1x_3 + c_7x_1 + c_8x_2 + c_9x_3 + c_{10},
\end{align}
and $\mathbf{c} \in \mathbb{R}^{10}$ denotes the parameters of the
quadric and $\mathbf{x} \in \mathbb{R}^3$ denotes the Cartesian
coordinates of a point on the surface. In principle, we would like to
solve for the parameters that minimize the sum of squared geometric
distances between the points and the quadratic surface. Unfortunately,
it turns out that this is a non-convex optimization problem with no
known analytical solutions. Instead, a standard approach is to solve
for an {\em algebraic fit}, that is to solve for the parameters
$\mathbf{c}$ that minimize
\begin{equation}
\sum_{i=1}^n f(\mathbf{c},\mathbf{x}^i)^2 = 
\mathbf{c}^T M \mathbf{c},
\label{eqn:algebraic_fit}
\end{equation}
where $M = \sum_{i=1}^n l(\mathbf{x}^i) l(\mathbf{x}^i)^T$,
$\mathbf{x}^1, \dots, \mathbf{x}^n \in R^3$ are the points to which
the curve is fitted, and 
\[
l(\mathbf{x}) = (x_1^2, x_2^2, x_3^2, x_1x_2, x_1x_3, x_2x_3, x_1, x_2, x_3, 1)^T.
\]
This problem is slightly different from the geometric fitting problem
because $f(\mathbf{c},\mathbf{x}^i)$ is not linear in the distance to
the surface. However, $f(\mathbf{c},\mathbf{x}^i)$ {\em is} a good
approximation to the geometric distance within a neighborhood about
the surface, and as a result, this general approach can yield good
fits.

\begin{figure}[t]
\begin{center}
    \subfigure[Quadric fit to a curved surface]{\includegraphics[height=1.0in]{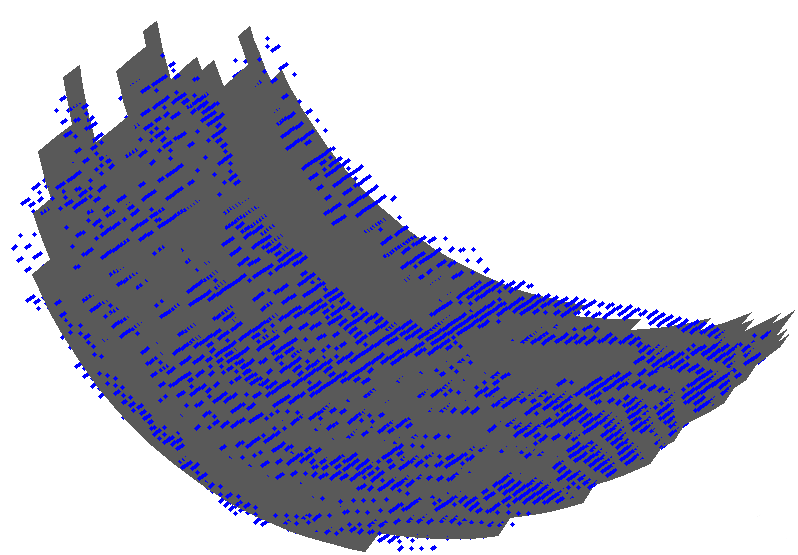}}
    \subfigure[Quadric fit to a corner surface]{\includegraphics[height=1.0in]{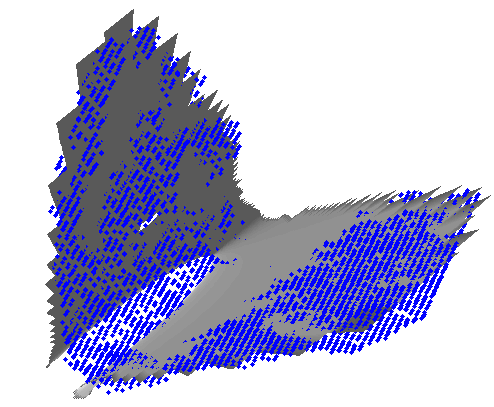}}
\end{center}
\caption{Examples quadric fits found using Taubin's method. Notice that a curved surface can be fit well by an ellipsoidal quadric whereas a corner surface can be fit well by a hyperbolic quadric. Each of these fits was calculated efficiently by solving a single $10 \times 10$ generalized Eigenvalue problem.}
\label{fig:taubinfit}
\end{figure}

An important choice that affects the resulting algebraic fit is how
Equation~\ref{eqn:algebraic_fit} is normalized. Notice that minimizing
Equation~\ref{eqn:algebraic_fit} directly would yield the trivial
solution with $\mathbf{c}$ at the origin. This question has been
studied extensively in the literature. Perhaps the simplest solution
is to constrain $\|c\|^2 = 1$. Then, Equation~\ref{eqn:algebraic_fit}
can be minimized by performing an Eigen decomposition on $M$. Other
possible solutions include setting $c_{10} = 1$~\cite{rosin_prl1993}
and setting the constraint $c_1^2 + \frac{1}{2}c_2^2 + c_3^2 =
1$~\cite{bookstein_cgip1979}. Unfortunately, both of the above
normalization methods can cause the algebraic fit to diverge
significantly from the solution to the geometric least squares problem
and produce poor fits. One normalization method that has been found to
work well in practice~\cite{andrews_cad2013} is Taubin's
method~\cite{taubin_pami1991}. This method sets the constraint
$\|\nabla_\mathbf{x} f(\mathbf{c},\mathbf{x}^i) \|^2 = 1$ and can be
solved as follows. Let
\[
N = \sum_{i=0}^n 
l_x(\mathbf{x}^i) l_x(\mathbf{x}^i)^T + 
l_y(\mathbf{x}^i) l_y(\mathbf{x}^i)^T + 
l_z(\mathbf{x}^i) l_z(\mathbf{x}^i)^T,
\]
where $l_x(\mathbf{x})$ denotes the derivative of $l(\mathbf{x})$
taken with respect to $x_1$ and the other derivatives are defined
similarly. Then, solve the generalized the generalized Eigen
decomposition, $\left(M - \lambda N \right) \mathbf{c} = 0$. The
best-fit parameter vector is equal to the Eigenvector corresponding to the
smallest Eigenvalue. Figure~\ref{fig:taubinfit} shows two examples of
surface fits that were found using Taubin's method. The point data in
these two examples come from a 3-D point cloud measured using an Asus
Xtion Pro sensor. Figure~\ref{fig:taubinfit}(a) shows a section of a
quadric fit to a set of points that lie on the side of a
cylinder. Figure~\ref{fig:taubinfit}(b) shows a section of a fit to
points on a right angle corner. The ability of the Taubin fit to
measure this kind of local neighborhood surface geometry makes it a
good tool for detecting target geometries in point clouds.

\subsection{Identifying Candidate Grasp Affordances}

Quadric fitting is a convenient method for efficiently finding object
surfaces that could potentially fit within the capture radius of the
robot hand and thereby satisfy the first EGA condition
(Section~\ref{sect:approach}). Although the object surface may be
non-convex, there must be a smoothed version of that surface that is
convex and is sufficiently curved in order to fit inside the capture
radius. The fitted quadric can be used to detect this condition
because it smooths out high-frequency content in the points to which
it has been fit. Given a quadric that has been fit to a point
neighborhood, we can evaluate the general shape of the neighborhood by
looking at the curvature of the quadric. A sufficiently large
curvature over much of the quadric indicates that it could potentially
fit within the capture radius.

Before proceeding, it is necessary to understand some details about
calculating the curvature of an implicit quadric. The curvature of a
quadratic surface at a particular point can be calculated by
evaluating the shape operator\footnote{In general, the shape operator,
  $S$, can be calculated using the first and second fundamental forms
  of differential geometry: $S = \mathbf{I}^{-1} \mathbf{II}$.} on the
plane tangent to the point of interest. The Eigenvectors of the shape
operator describe the principle directions of the surface and its
Eigenvalues describe the curvature in those directions. This can be
calculated for a point, $\mathbf{x}$, on the surface by taking the
Eigenvalues and Eigenvectors of:
\[
\left(I - N(\mathbf{x})N(\mathbf{x})^T\right)\nabla N(\mathbf{x}),
\]
where $N(\mathbf{x})$ denotes the surface normals of the quadric. It
is calculated by differentiating and normalizing the implicit surface:
\[
N(\mathbf{x}) = \frac{\nabla f(\mathbf{c},\mathbf{x})}{\|\nabla f(\mathbf{c},\mathbf{x})\|},
\]
where 
\[
\nabla f(\mathbf{c},\mathbf{x}) = 
\left(
\begin{array}{c}
2c_1x_1 + c_4x_2 + c_6x_3 + c_7 \\
2c_2x_2 + c_4x_1 + c_5x_3 + c_8 \\
2c_3x_3 + c_5x_2 + c_6x_1 + c_9
\end{array}
\right).
\]

Once a quadric is fit to a point neighborhood, we evaluate the median
curvature of the quadric in the point neighborhood. This is
accomplished by randomly sampling several points from the local
quadric surface and calculating the maximum curvature (maximum of the
two principle curvatures) magnitude at each of them. Then, we take the
median of these maximum curvature values and accept as grasp
affordance candidates all quadrics where the median curvature is
larger than that implied by the hand capture radius. This method
detects smoothly curving surfaces such as that shown in
Figure~\ref{fig:taubinfit}(a), where the majority of sampled
neighborhood points are sufficiently curved.

\section{Grasp Affordance Perception Pipeline}

\begin{algorithm}
\caption{Grasp Affordance Perception}
\begin{algorithmic}[1]
\Procedure{AffordanceFind}{$\mathbf{p}_{cloud}$, $n$, $r_{target}$} 
   \State $list = \{\}$;
   \For{$i=1:n$} 
     \State sample point neighborhood, $\mathbf{p}_i$
      \If{$!FilterOcclusion(\mathbf{p}_i)$}
      \State $\kappa_{max}, \nu_{axis} = FitTaubin(\mathbf{p}_i)$; 
      \If{$\kappa_{max} \geq 1/r_{target}$} 
      \State $found, \lambda_{shell} = FitShell(\mathbf{p}_i,\nu_{axis})$;
      \If{$found$} 
      \State $list \leftarrow \{list, \lambda_{shell}\}$;
      \EndIf
      \EndIf
      \EndIf
   \EndFor
\EndProcedure
\end{algorithmic}
\end{algorithm}

The grasp affordance perception pipeline works by sampling a large set
of neighborhoods from the point cloud, identifying grasp affordance
candidates using Taubin fitting, and then attempting to fit a
cylindrical shell with the appropriate capture radius and
thickness. Pseudocode is shown in Algorithm 1.

The algorithm works as follows. Step 4 samples neighborhoods from the
point cloud. For each point neighborhood, Step 5 eliminates from
consideration those neighborhoods that are significantly occluded by
objects in the foreground. Step 6 does Taubin quadric fitting and Step
7 filters out those neighborhoods that are not sufficiently curved to
satisfy the first EGA condition. For the neighborhoods that remain,
Step 8 fits a cylindrical shell in the neighborhood that satisfies the
second EGA condition. These steps are discussed in more detail below.

\subsection{Sampling and Occlusion Filtering}

Step 4 samples point neighborhoods by randomly sampling a single point
from the cloud and setting the point neighborhood equal to the set of
points that fall within a sphere with radius equal to the capture
radius. This value of the radius ensures that the point neighborhoods
are roughly proportional to the size of the robot hand.

Step 5 eliminates from consideration point neighborhoods that are
significantly occluded by points in the foreground. This is an
important step because when a point cloud is constructed from a single
range image, occlusions can introduce significant ambiguity into
background parts of the cloud. If the foreground shadows items in the
background, shapes can appear in the background that do not really
exist. This is a particular problem for our grasp affordance detector
because foreground shapes can easily introduce shadows that cause the
EGA conditions to be erroneously satisfied. Fortunately, this kind of
occlusion is easily detected. For each neighborhood, we project the
sphere that defines the neighborhood onto the range image (forming a
circle in the range image). We take all points within this circle and
evaluate their range. If more than a threshold number of these points
are closer than the closest point in the neighborhood, then we assume
that we have detected a potential occlusion and discard the
neighborhood. When a point cloud is constructed by registering data
from two or more range images together, it is more difficult to use
the above method to identify occlusions. A simple extension is to
label a point neighborhood as unoccluded when the neighborhood is not
occluded in any range image. However, we have not tested this
extension.

\subsection{Fitting the Cylindrical Shell}

After doing Taubin fitting (Step 6) and filtering out neighborhoods
without a sufficiently large curvature (Step 7), we fit the
cylindrical shell. We are searching for a shell that contains a large
number of points inside the inner radius but contains very few points
within the thickness of the shell itself. Unfortunately, fitting a
cylindrical shell as described above is a non-convex problem and
cannot be solved directly using regression. Moreover, a brute force
search in the five-dimensional search space (four dimensions of pose
plus one dimension of radius) is computationally too
expensive. Instead, we reduce the size of the search space by setting
the axis of the cylindrical shell to be equal to the axis of maximum
curvature found during Taubin fitting. This reduces the search space
from five dimensions down to three (two position dimensions and one
radius). Since even a three-dimensional search is prohibitive when
executed for a large number of candidate neighborhoods, we simplify
the search further by: 1) performing a cylinder fitting step to
establish a centroid for the inner cylinder of the shell, and 2)
searching over the 1-D space of shell radii for the smallest radius
that satisfies the EGA conditions.

\subsubsection{Cylinder Fitting}

The purpose of doing the cylinder fit is to find the shell
centroid. We assume that the cylinder fit will find a close
approximation to the inner cylinder of the shell and that we will be
able to find a good shell fit by subsequently increasing the shell
radius. In order to perform the cylinder fit, we first set the
orientation of the axis of the cylinder to that of the axis of maximum
curvature at the median point on the fitted quadric (see
Section~\ref{sect:taubinfit}). Once the cylinder axis is fixed, we can
calculate the closest fitting cylinder by projecting the point
neighborhood onto the plane orthogonal to the axis and finding the
best-fit circle. Let $W =(w_1,w_2) \in \mathbb{R}^{3\times2}$ be a
basis for the orthogonal plane. Then the projection of point
$\mathbf{x}^i$ onto the plane is calculated: $\mathbf{\bar{x}}^i = W^T
\mathbf{x}^i$, where $\mathbf{\bar{x}}^i = (\bar{x}^i,
\bar{y}^i)^T$. As was done with the quadric fitting, we calculate the
best-fit circle by using algebraic distance. We want to find the
parameters, $h_x$, $h_y$, and $r$ that minimize:
\begin{align} \nonumber
\sum_{i=1}^n ( (\bar{x}^i - h_x)^2 + & (\bar{y}^i - h_y)^2 - r^2 )^2 \\
& = \sum_{i=1}^n ( (\bar{x}^i)^2 + (\bar{y}^i)^2 + a\bar{x}^i + b\bar{y}^i + c )^2,
\label{eqn:algebraic_circle}
\end{align}
where $a = -2h_x$, $b = -2h_y$, and $c = h_x^2 + h_y^2 -
r^2$. Equation~\ref{eqn:algebraic_circle} can be solved for
$\mathbf{w} = (a, b, c)^T$ using standard calculus. The result is:
\begin{equation}
\mathbf{w} = - 
\left( \sum_{i=1}^n \mathbf{l}_i \mathbf{l}_i^T \right)^{-1}
\sum_{i=1}^n  \lambda_i \mathbf{l}_i,
\label{eqn:cicle_fit}
\end{equation}
where $\lambda_i = (\bar{x}^i)^2 + (\bar{y}^i)^2$ and $\mathbf{l}_i =
(-\bar{x}^i, -\bar{y}^i, 1)^T$. After solving
Equation~\ref{eqn:cicle_fit} for $\mathbf{w}$, we back out the circle
center, $(h_x, h_y)$, and radius, $r$.

\subsubsection{Finding the Shell}

\begin{figure}
\begin{center}
    \subfigure[]{\includegraphics[height=1.4in]{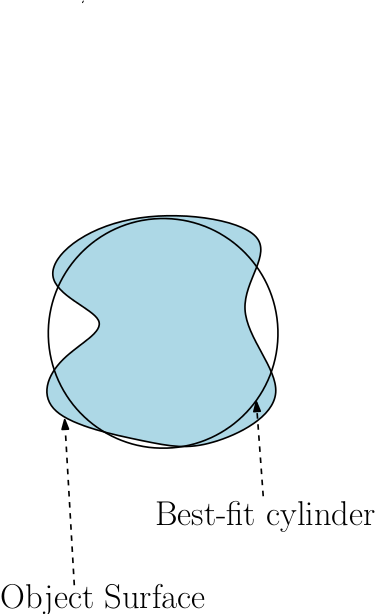}}
    \subfigure[]{\includegraphics[height=1.4in]{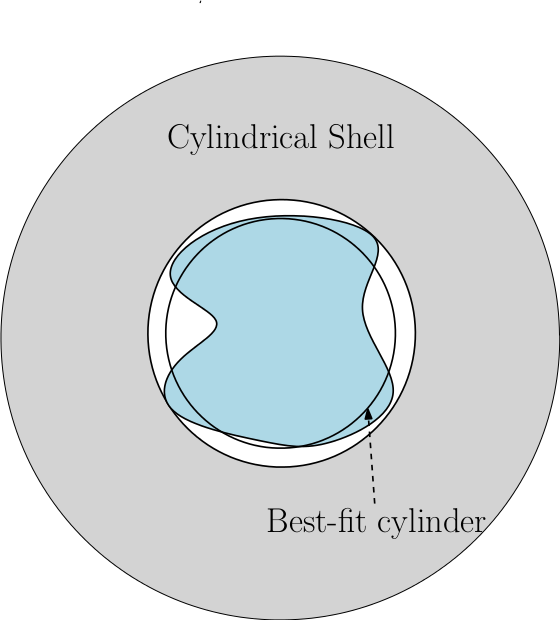}}
\end{center}
\caption{(a) shows the cylinder fitted to the neighborhood points on
  the object surface. (b) shows the cylindrical shell found by
  increasing the radius until the fitted neighborhood points are
  contained within the inner radius.}
\label{fig:capturing_cylinder}
\end{figure}

After fitting the cylinder to the point neighborhood, we seed the
search for the cylindrical shell by setting the position and
orientation of the shell axis equal to that of the fitted cylinder
axis. The only remaining unknown shell parameter is the
radius. Starting with the radius of the fitted cylinder, we
iteratively increase the shell radius in small steps while keeping the
shell thickness constant and equal to the finger thickness. We
increase the shell radius until we find a radius where few or no
points are contained within the shell thickness. This process is
illustrated in
Figure~\ref{fig:capturing_cylinder}. Figure~\ref{fig:capturing_cylinder}(a)
shows the best fit cylinder. Figure~\ref{fig:capturing_cylinder}(b)
shows the cylindrical shell found by increasing the cylinder radius
until the affordance surface is contained.

\section{Experiments}
\label{sect:experiments}

Figure~\ref{fig:front_example} illustrates the typical performance of
our approach. In this example, we assume that the robot has a maximum
capture radius of $2.9$ cm. At this radius, there are seven enveloping
grasp affordances (the apple, the end of the banana, the neck of the
squirt bottle, the dustpan handle, the jug handle, the jug cap, and
the broom handle) that could potentially be grasped by an enveloping
robot hand. Figure~\ref{fig:front_example}(a) shows the affordances
circled manually in an RGB image. Figure~\ref{fig:front_example}(b)
shows the EGA geometries found automatically by our algorithm. Notice
that there is an exact correspondence between the affordances found in
the two images. In addition to evaluating the absolute precision and
recall of our method, we compare our approach to two possible
alternative algorithms. To our knowledge, there are no other
algorithms in the literature that address the grasp affordance
localization problem in a way that can easily be compared to our
work. Therefore, we propose two variations on our algorithm that
replace our use of Taubin's method with alternative shape estimation
techniques that are often found in the 3-D point cloud literature. Our
results indicate that Taubin-based fitting does have better recall
(with the same precision) than the alternative methods, but that all
the methods we considered {\em can} have very good performance on some
datasets.

\begin{figure*}[t]
\begin{center}
    \subfigure[Mugs]{\includegraphics[height=1.1in]{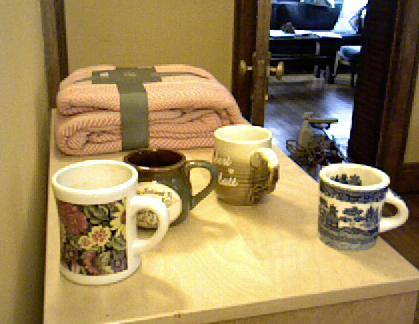}}
    \subfigure[Cleaning1]{\includegraphics[height=1.1in]{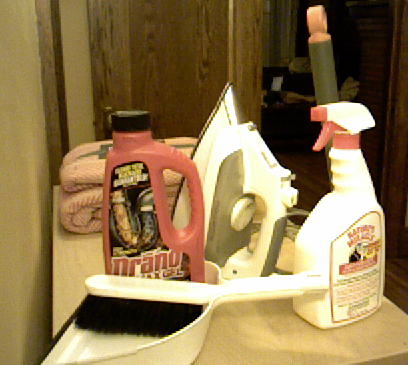}}
    \subfigure[Kitchen1]{\includegraphics[height=1.1in]{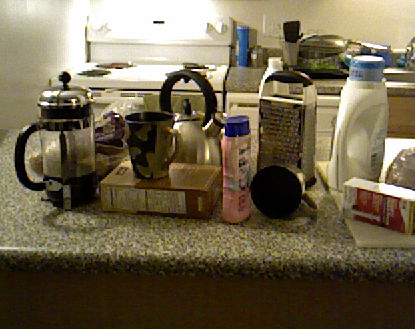}}
    \subfigure[Cleaning2]{\includegraphics[height=1.1in]{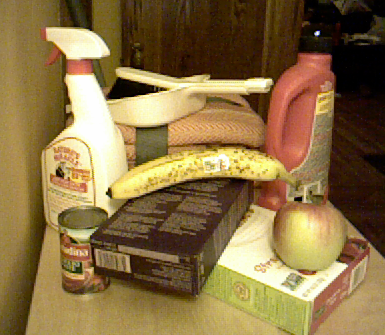}}
    \subfigure[Household1]{\includegraphics[height=1.1in]{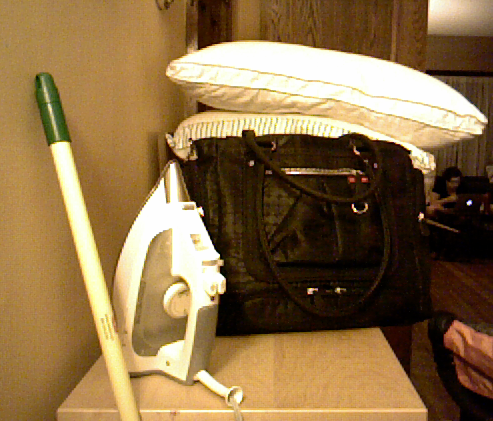}}
\\
    \subfigure[Household2]{\includegraphics[height=0.95in]{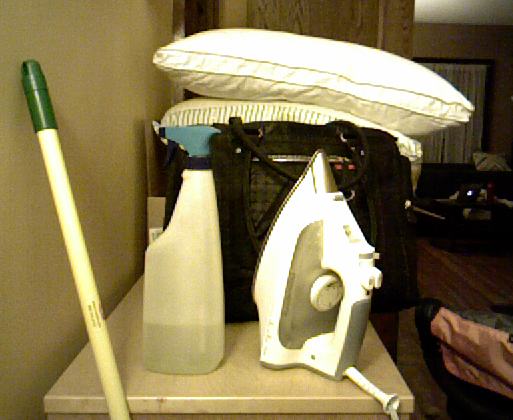}}
    \subfigure[Kitchen2]{\includegraphics[height=0.95in]{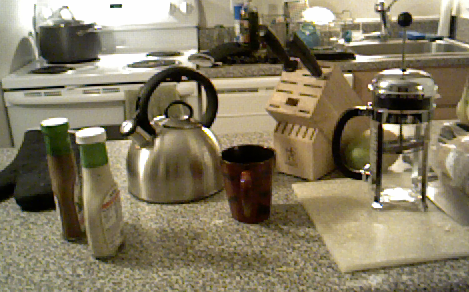}}
    \subfigure[Kitchen3]{\includegraphics[height=0.95in]{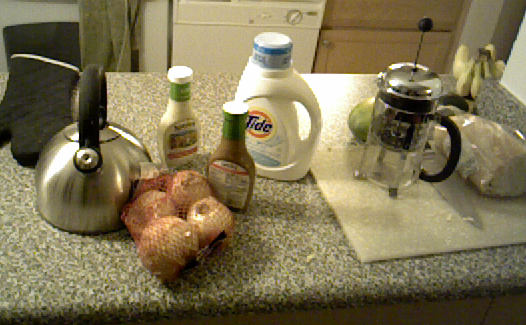}}
    \subfigure[Luggage]{\includegraphics[height=0.95in]{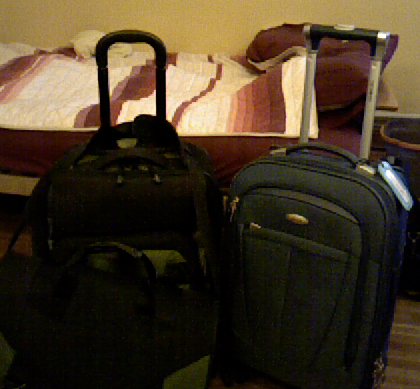}}
    \subfigure[Cleaning3]{\includegraphics[height=0.95in]{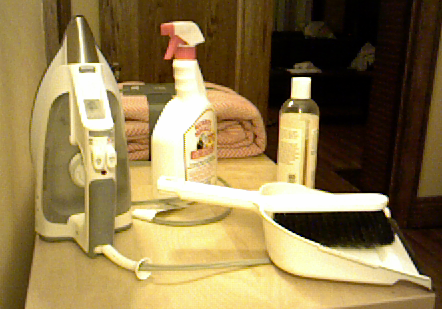}}

\end{center}
\caption{RGB images corresponding to the ten point clouds used in our
  experiments (all data collected using the ASUS XTion Pro). Our
  algorithm detects and localizes the enveloping grasp affordances in
  these scenes.}
\label{fig:data_sets}
\end{figure*}

\subsection{Comparisons}

The key feature of our proposed algorithm is the Taubin quadric
fitting in Step 6 of Algorithm 1. This step does two things: it
enables the algorithm to filter out low-curvature neighborhoods from
further consideration and it enables us to decrease the dimension of
the prismatic annulus search space by fixing the axis of curvature. In
order to evaluate the significance of this step, we compare the Taubin
fit version of the algorithm with two variations on the algorithm.

\subsubsection{The PCA Variation}

The first alternative is to do standard PCA on each point neighborhood
instead of doing the Taubin fit: calculate the $3 \times 3$ covariance
matrix for the points in the neighborhood and perform an Eigen
decomposition. In this scenario, ``curvature'' of the point
neighborhood would be approximated (in some sense) by taking the ratio
of the second and third smallest Eigenvalues of the covariance
matrix. The axis of curvature would be calculated by taking the
Eigenvector associated with the largest Eigenvalue of the covariance
matrix. Unfortunately, we were unable to improve overall localization
performance by placing any threshold on the ratio of
Eigenvalues. Therefore, in this alternative algorithm scenario, we
omitted the filtering step (Step 7 of Algorithm 1) entirely and just
take the direction of the principle Eigenvector as the axis of
curvature. The rest of the algorithm is the same as shown in Algorithm
1. We will refer to this variation on our algorithm as the ``PCA
variation''.

\subsubsection{The Normals Variation}

The second alternative is to calculate a covariance matrix on surface
normals for points within each point neighborhood rather than doing
the Taubin fit. In this scenario, we assume that the point cloud data
has been pre-processed by estimating the surface normal for each point
in the cloud using PCA on a 3cm radius about each point. Then, for
each point neighborhood, we calculate the $3 \times 3$ covariance
matrix, $M = \sum_{i=1}^k \mathbf{n}_i \mathbf{n}_i^T$, where
$\mathbf{n}_1, \dots, \mathbf{n}_k$ describe the surface normals for
points in the neighborhood. The ``curvature'' of the point
neighborhood can be estimated by taking the ratio of the second
largest and the largest Eigenvalues of $M$. The axis of curvature for
the neighborhood can be estimated by taking the Eigenvector
corresponding to the smallest Eigenvalue. As before, we were unable to
improve algorithm performance by doing any thresholding on this
value. Instead, this alternative algorithm omits curvature filtering
(Step 7, Algorithm 1) and sets the axis of curvature as described
above. We will refer to this variation on our algorithm as the
``Normals variation''.

\subsection{Methods and Results}

\begin{figure}
\begin{center}
    \includegraphics[width=3.75in]{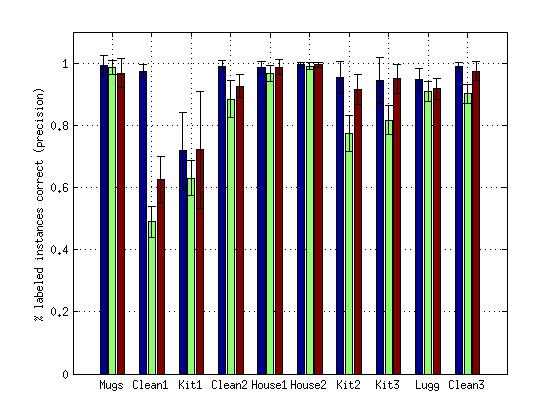}
\end{center}
\caption{Precision of each of the three methods for the ten datasets
  averaged over ten runs each. For each dataset, Taubin performance is
  shown in blue (leftmost bar in each triple), PCA performance is
  shown in green (middle bar in each triple), and Normals performance
  is shown in red (rightmost bar in each triple. The error bars show a
  95\% confidence interval over the ten runs.}
\label{fig:results}
\end{figure}
To our knowledge, there are no datasets in the literature designed to
test enveloping grasp affordance detection. Therefore, we obtained a 
dataset of our own. Each of the scenes in our dataset is a point cloud created
using a range image captured using an Asus XTion Pro
sensor. Figure~\ref{fig:data_sets}(a -- e) shows the RGB images
associated with the data sets. The points in each point cloud 
corresponding to an enveloping grasp affordance (for the most part,
handles in the scene) were manually labeled. We labelled every surface
in each scene that had a radius smaller than the capture radius
(2.6cm), had sufficient clearance around it (at least 0.8cm), and was
at least three centemeters long. Many of the labelled enveloping grasp
affordances were handles (such as the handle on the top of the
tean-kettle in {\em Kitchen2}) but some of them were ``handle-like''
surfaces on objects (such as the topics of the salad dressing bottles
in {\em Kitchen2}.

The algorithm that was executed is exactly as it appears in Algorithm
1. The point clouds in our dataset were each comprised of
approximately 250k valid depth points. In general, our algorithm does
not require any pre-processing of the point cloud. There is no
voxelization step. We did a surface normals calculation step (with a
3cm neighborhood size) only for the purpose of implementing the
Normals Variation version of our algorithm. There is no ground plane
separation step. There is no object segmentation required. There are
no assumptions about objects pointing in the direction of the gravity
vector. In all of our experiments, we parameterized the algorithm with
a capture radius of 2.6cm and a required shell gap of 0.8cm. Each run
of the algorithm sampled the cloud 4000 times ({\em i.e.} the value
for $n$ in step 3 of Algorithm 1 was 4000).

Figure~\ref{fig:results} illustrates the results of our
comparison. For each of the ten scenes, we ran each of the three
algorithms and averaged the results over ten different runs. The
recall (percent of ground truth affordances found by the algorithm)
was 100\% on nearly every run for every algorithm, so we do not report
that result. However, precision (percent of labeled affordances that
were correct) varied significantly. Perhaps the most striking result
is that all of the algorithms performed very well on many of the
datasets. On many of the scenes, the Taubin method had very similar
precision to the Normals method (at least 90\% on eight out of ten
datasets). In nearly all the scenes, the PCA method performed slightly
worse. Two of the scenes deserve particular attention. First, notice
that while the Taubin method performed well on {\em Cleaning1}, PCA
and Normals did not. The reason for this difference is that PCA and
Normals both detected the side of the dust pan as a handle whereas the
Taubin method did not because it was not sufficiently curved. Second,
notice that all three methods perform relatively badly on the {\em
  Kitchen1} scene. This is a result of false handle detections on the
side of the lotion bottle in the middle of the scene, on the
horizontal box at the right, and inside the glass French press at the
left of the scene.

\subsection{Practical Running Time}

\begin{figure}
\begin{center}
    \includegraphics[width=3.75in]{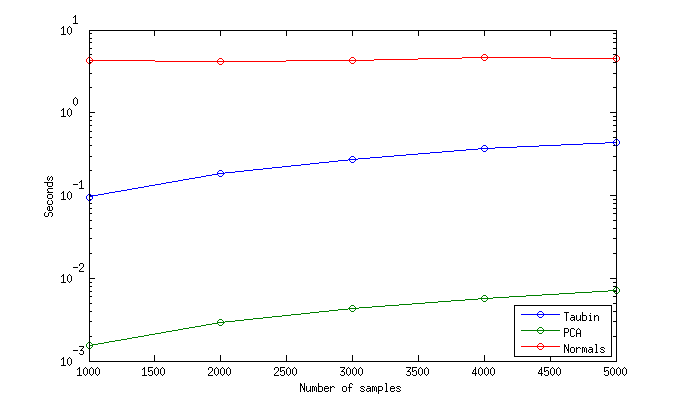}
\end{center}
\caption{Average runtime of each of the three algorithm variations over 10 runs.}
\label{fig:results_time}
\end{figure}

In order to evaluate whether the algorithm is efficient enough to be employed on a real robotic system, we implemented the algorithm in C++ for the robot operating system ROS, and compare the practical runtime of the three variations of the algorithm. The code is parallelized, and a k-nearest neighbor search is used instead of the spherical radius search ($k = 500$). This experiment was performed on a system with 6GB of system memory and an Intel i7 CPU with 2GHz and four physical CPU cores.

Figure~\ref{fig:results_time} shows the results of our comparison. PCA turns out to be the fastest of the three variations, requiring less than $10\%$ of the time that Taubin consumes, and even staying below $0.01$sec for $5000$ samples. Taubin is the second fastest, and its runtime increases linearly with the number of samples. The main reason for the time difference between PCA and Taubin is that the PCA variation only requires to calculate a 3x3 covariance matrix and to solve the eigenvalue problem for this matrix, while the Taubin variation requires to calculate two 10x10 matrices and to solve a generalized eigenvalue problem for these two matrices. The Normals variation is the slowest, and consumes about $4$sec. The reason for this is that it needs to calculate the surface normals for each point in the point cloud.

The low time consumption of the PCA and the Taubin variation emphasizes the practical applicability of the algorithm to the perception-for-grasping problem on a real robotic system.

\section{conclusion}

The main contribution of this paper is a perception pipeline that solves the grasp-perception problem in a robust and computationally cheap way. Our method works well for localizing enveloping grasp affordances in cluttered environments, and it is fast enough to be applied to real-world scenarios.

The algorithm presented in this paper can be applied "out of the box" to a supervised autonomy scenario in which a
human operator selects which grasp affordance is to be grasped by the robot. By virtue of selecting the affordance, the human has a great deal of control over exactly how the robot will perform the grasp. 

Other applications are completely autonomous scenarios where the robot itself needs to decide where to grasp. Grasp affordances allow to decouple the process of controlling the robot's actions from the grasp planning process.

The next step for our work is to use the grasp affordances provided by our algorithm for grasp planning. Our vision is to implement this algorithm on a robot and to use it in a real-world grasping application.

\bibliographystyle{IEEEtran}

\bibliography{platt}

\end{document}